# A Directed-Evolution Method for Sparsification and Compression of Neural Networks with Application to Object Identification and Segmentation – and considerations of optimal quantization using small number of bits

Luiz M Franca-Neto

(Abstract) In seeking effective neural network models for use in power and resource constrained edge computing, model pruning (forced sparsification) and optimal parameter representation with diminished number of bits have become central strategies for successful deployment. Pruning became especially attractive when it was shown that a pruned large neural network outperforms a "dense" smaller network with the same memory footprint [1]. Pruning methods were proposed under the assumption that the magnitudes of the network parameters are good proxies for the relevance of those parameters to the neural network's accuracy [2,3]. Hence, these methods attempt to zero low magnitude parameters. Pruning methods based on Hessian approximations (curvature information) around network's loss minima were reported to consistently outperform those methods based solely on parameters' magnitudes [4]. Both methods were shown to be superior to merely random methods since these zero parameters with no previous consideration of their impact in network accuracy [5]. These results combined appear to suggest that the first two approaches are using proxies for the relevance of parameters and that there is space for improving on those results further if the actual relevance of the parameters is more accurately determined before zeroing them. This work introduces Directed-Evolution (DE) method for sparsification of neural networks, where the relevance of parameters to the network's accuracy is directly assessed and the parameters that produce the least effect on accuracy when tentatively zeroed are indeed zeroed. DE method avoids a potentially combinatorial explosion of all possible candidate sets of parameters to be zeroed in large networks by mimicking evolution in the natural world. DE starts at a pre-trained network and progressively "evolves" towards a much sparser network model using cycles of *cumulative small steps* of minimal loss of performance and re-training. In addition, DE uses a distillation context [5]. In this context, the original network is the "teacher" and DE evolves the "student" neural network to the sparsification goal while maintaining minimal divergence between teacher and student. The DE's distillation context provides three benefits: (a) the student network captures relevant knowledge learned by the pre-trained "teacher" network, (b) the degree to which the teacher network is over-parameterized is made clear at every cycle of the DE method and (c) the dataset used in the retraining stage in each DE cycle can be much smaller than the dataset previously used in the training of the teacher network. After the desired sparsification level is reached in each layer of the network by DE, a variety of quantization alternatives are used on the surviving parameters to find the lowest number of bits for their representation with acceptable loss of accuracy. A procedure to find optimal distribution of quantization levels in each sparsified layer is presented. Suitable final lossless encoding of the surviving quantized parameters is used for the final parameter representation. DE was used in sample of representative neural networks using MNIST, Fashion-MNIST and COCO data sets with progressive larger networks. An 80 classes YOLOv3 with more than 60 million parameters network trained on COCO dataset reached 90% sparsification and correctly identifies and segments all objects identified by the original network with more than 80% confidence while using 4-bit parameter quantization. DE produced final memory footprint compression between 40x and 80x in the examples studied. It has not escaped the authors that techniques from different methods can be nested, and that once the best parameter set for sparsification is identified in a cycle of DE, a decision on zeroing only a sub-set of those parameters can be made using a combination of criteria like parameter magnitude and Hessian approximations. In this current paper, only results for the case of purely DE method are reported, and results for nesting of methods inside DE will be reported in a follow-up paper.

Author contact: franca@ieee.org

## 1. INTRODUCTION

The design of very low power machine learning inference-only accelerators aims at minimizing unnecessary movement parameters and data used by a neural network during computation [8]. For that effect, weight stationarity and placement of all the necessary parameters in a single die are commonly used in always-on battery powered solutions [9]. Fitting all the parameters on a single die however becomes more challenging as morse sophisticated Artificial Intelligence applications at the edge drive the need for neural networks using between 10 million and 100 million parameters in the near future. Hence, compressed representations of those large parameters set with minimal loss of neural network accuracy becomes an enabling feature for future low power machine learning accelerators.

It's frequently thought that any neural network might be plagued by a massive number of redundant parameters [1]. This can be caused by either (i) an over parameterized neural network for a given application or (ii) the original pre-trained neural network became over parameterized for a new specialized objective. Independent of the cause, a suitable method to find those redundant parameters and make them zero-valued parameters becomes a desired goal. In pursuit of this goal, three constituent challenges need to be addressed:

(i) *find an effective computational procedure to properly identify the redundant parameters that can be zero-valued, effectively removing their corresponding edges from a fully connected neural network or removing their participation in kernels of convolution layers, with minimal or acceptable loss in the network accuracy performance,*

(ii) *represent the surviving parameters with the smallest number of bits, also with minimal or acceptable loss in the network performance,*

*and*

(iii) *find a convenient coding scheme for properly*

*representing the surviving and quantized parameters in a compressed form. In this case the very distribution of values in the surviving parameters, i.e. their entropy, is analyzed for optimal representation of weights and use of short codewords for the most frequent values used by the neural network.*

The steps (i) and (ii) are therefore lossy steps. Step (iii) is lossless and Huffman coding frequently used in it. The main motivation for pursuing these 3-stepped approach is that sparsefied large networks consistently outperforms "dense" networks with the same parameter memory footprint [1-3].

This work tries also to include two additional soft constraints while sparsifying an original large network:

*(1) the original dataset used in training the large network may not be available or, by the large size of the original dataset, it is not desirable to be used. A much smaller suitable dataset is to be used for the sparsification.*

*(2) the sparsified network has objectives that are a subset of the original objectives used in training the original larger network*

The inclusion of (1) stems from the desirability of need faster re-training during the cycles of Directed-Evolution method and distillation discussed later. The inclusion of (2) is motivated by the fact that machine learning at the edge tend to involve a more specialized set of objectives. In case of object detection at the edge, the focus might be home security or navigation on city streets. In such a case, there might be higher emphasis on identifying people, domestic animals and vehicles than in identifying a plane or an eagle, for instance. The inclusion of (2) also tends to promptly generate an expectation of augment of redundancy in the original large network and its manifestation is discussed along the results of this work in session 4 of this paper.

This paper focus on compression per se and it is possible that the highest compression achievable places a cost on latency for each inference in neural network. The paper is organized with session 2 presenting the conceptual description of Directed-Evolution method *vis-à-vis* related works. Session 3 describes preliminary studies of Directed-Evolution applied to classification tasks using the workhorse examples of MNIST and Fashion-MNIST. Session 4 illustrates the use of Directed-Evolution on a YOLOv3 neural network trained on COCO dataset and sparsified to 90%. Session 5 discusses the new vistas opened by Directed-Evolution method for futher work. Then we summarize our contributions.

## 2. THE DIRECTED-EVOLUTION METHOD AND RELATED WORKS

Popular approaches to sparsification (pruning) in machine learning assume that parameters' small maginitude values correlate with the parameter significance in the network accuracy [1-4]. Hence, those approaches attempt at zeroing the smallest valued parameters in a network. Other approaches include information on curvature on the minima of the loss function of a network for estimating the significance of parameters in a network. This curvature information is brought to these methods through Hessian approximations. In these methods, products of Hessian coefficients and parameters magnitude squared are taken to represent parameters' relevance to network accuracy. Those parameters with smaller values for that product are then zeroed [2-4].

It should be noticed that the first derivatives for each parameter at that final well trained model would be ideally zero (local minimum) for the loss function, and the second derivatives would provide information on the quality of that minimum, maximum or inflexion point. A parameter at a "shallow" minimum for the loss function can be *varied* more than a parameter at a "deep" minimum for the same effect in the loss function of the neural network, all other parameters being the same. Zeroing the parameters on shallow minima is the implicit motivation for the pruning methods using Hessian approximations. It's however less obvious that the parameters in shallow minima should be automatically the best candidates to be zeroed. It should also be mentioned that (as pointed by [4]) that if all the Hessian coefficients were equal, parameters would be zeroed according to their magnitudes as it is done in previous pruning methods – hence the superset nature of the Hessian approximation methods in pruning. The fact that these methods perform quite well and reach for levels of sparsity higher than 70% [1-5] suggests perhaps that their criteria for zeroing parameters are *good proxies* to the parameter's true relevance for the network accuracy. There is thus an opportunity to advance methods that go beyond proxies and more directly determine parameters' relevance before zeroing them.

Differently from previous approach, Directed-Evolution attempts at directly evaluate a parameter relevance to the neural network loss function. No proxy is used. In Directed-Evolution (DE), the significance of parameters is determined before they are zeroed.

In DE, the challenge than becomes how to effectively explore the space of parameters in a neural network to find those that can be zeroed. Unfortunately, the parameter space is too large and their potential interaction largely unpredictable for any hope of an exhaustive exploration. This combinatorial explosion can be appreciated by considering a relatively small layer with 1k parameters of which it would be desirable to find the 500 parameters set that, together, makes the smallest impact in performance accuracy if they were zeroed for 50% sparsity. Testing the impact in accuracy of all possible 500-member sets out of 1k parameters implies testing a sample space with size $\sim 2.7 \times 10^{299}$ . Such a sample space already surpasses in many orders of magnitude the number of atoms in the known universe ($\sim 10^{78}$). The challenge would only be more difficult with layers with more than 1 million parameters or with entire networks with more than 100 million parameters.

Directed-Evolution method improves on these combinatorial odds by using a procedure akin to Evolution in the natural world. Neither Evolution in the natural world has hopes of exploring all possible genetic variations in living organisms nor Directed-Evolution method has any hope of exploring all possible alternatives for parameter sparsification.

Evolution in the natural world progresses with very small cumulative steps from a viable organism. A random small number of mutations, sometimes single mutations even, generate variations of that organism. These variations are then exposed to the screening of the fittest (natural selection), which is a deterministic step, and the fittest of those variations generates larger number of descents. The process then repeats. The 3 stages over which Directed-Evolution cycles are described in figure 1 with the closely corresponding stages of Evolution in the natural world in parentheses. Directed-Evolution (DE) method is described in a context of distillation. starts from a pre-trained "teacher" network (the viable organism in the natural world). In stage 1, random trials of small set of parameters, even a single parameter sets, from a large layer, or from each layer in a set of layers, or from each layer in the whole neural network is chosen. In stage 2, the set from these random trials which produces the smallest divergence between teacher and student networks (equivalent to smallest impact in the accuracy of the network) if all the set members are zeroed is chosen and in effect zeroed. In stage 3, the network with the surviving parameters (the non-zeroed parameters) is retrained. The process then is repeated to cumulatively determine the next small set of parameter or even the single one parameter with the smallest impact in network accuracy all the way to a target sparsification level.

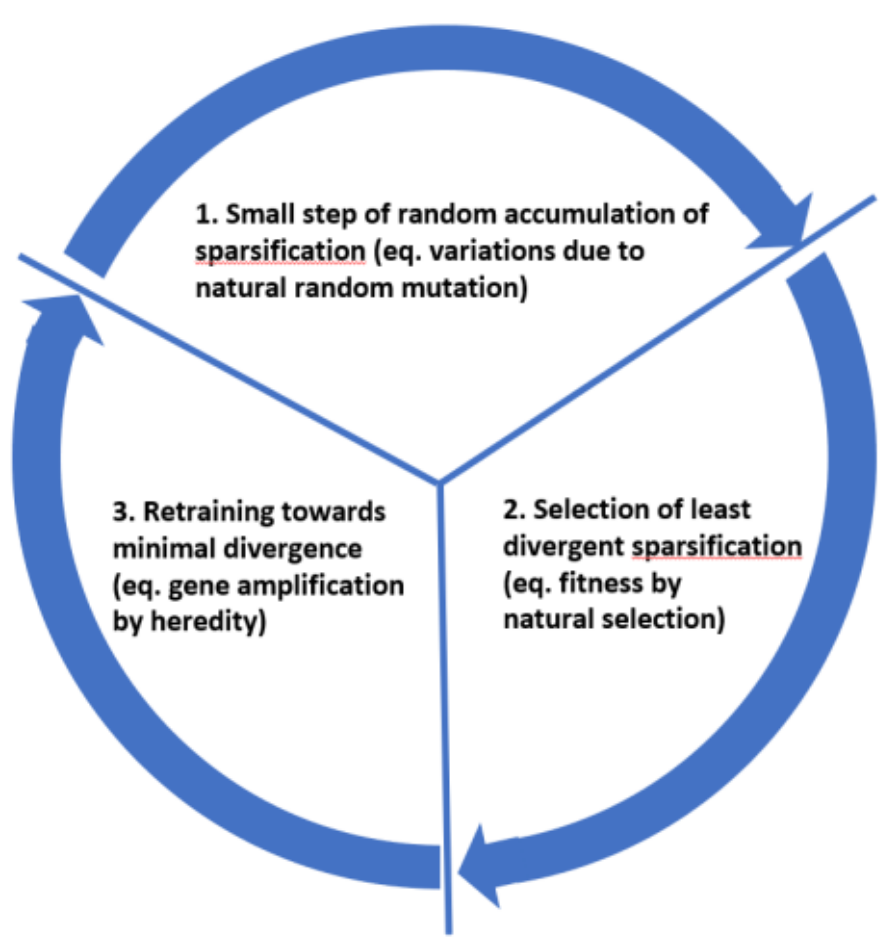

Figure 1: Directed-Evolution (DE) method repeat a cycle progressively augmenting the sparsification of a neural network. The DE cycle consist of three states: (1) a number of trials randomly generates candidate sets of parameters to be sparsified; (2) the candidate set of parameters which makes least divergence from a reference network when its members are zeroed is chosen and its parameters are irrevocably zeroed; and (3) the neural network with the newly zeroed parameters is retrained towards least divergence from the reference network. The cycle is repeated till the level of desired sparsification or a limit in the loss of performance or limit on divergence from the refence network is reached. In parentheses the corresponding stages in natural evolution.

Directed-Evolution (DE) method is therefore quite different from purely random methods, where weights are zeroed in single trials with no previous consideration of their impact in the network accuracy [5]. It is also a straightforward extension of the DE method to nest in stage 2 of its cycle considerations of weight magnitudes or Hessian approximations. That is, once the best set of parameters for zeroing is determined, instead of zeroing all the members of this set, considerations of magnitude or Hessian approximations can be used to zero only a sub-set of its members.

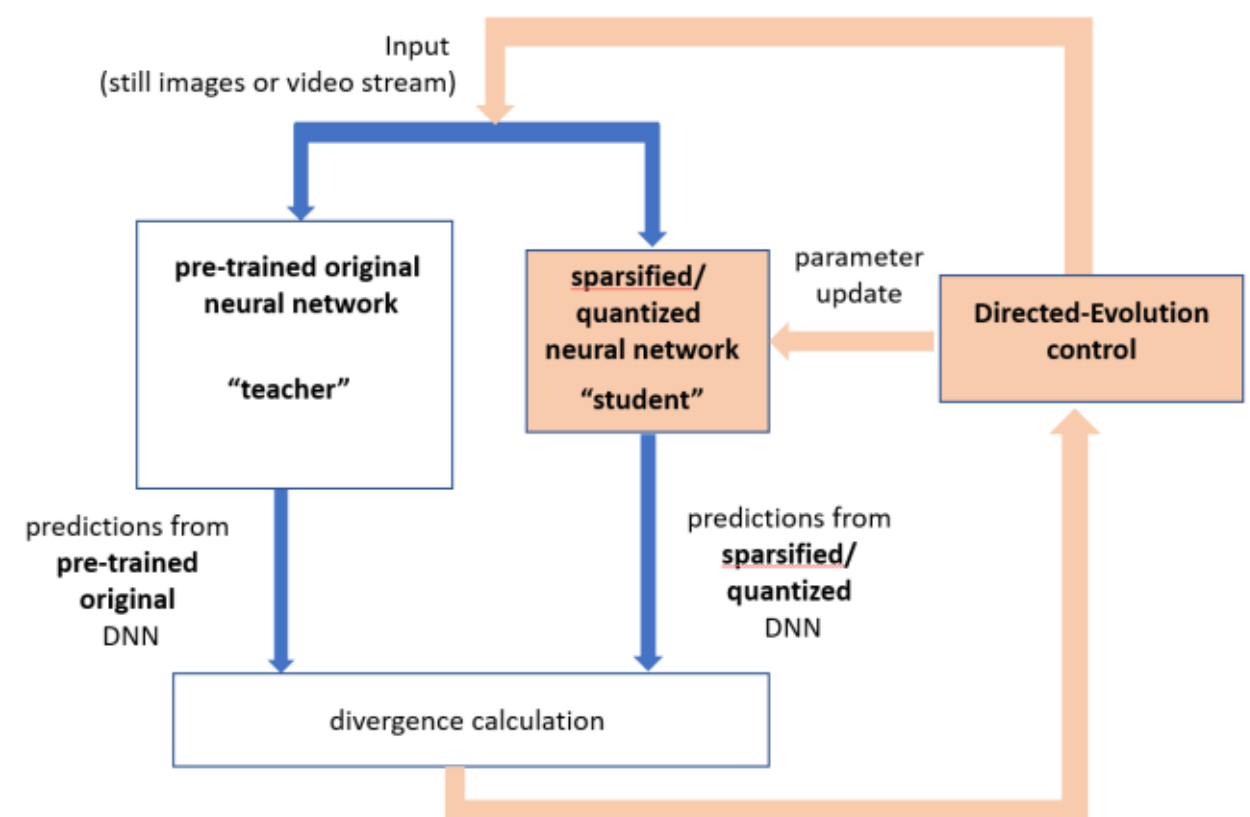

Figure 2: Distillation context of the Directed-Evolution (DE) method: a pre-trained large deep neural network, "teacher", and a copy neural network, "student", have their output compared as a measure of their dissimilitude. A Directed-Evolution control unit progressively make the student network more sparse while updating the parameters of the student network and monitoring the divergent measure of the teacher and student network outputs for inputs that are taken from a limited subset of the original training dataset used when training the teacher network.

It came to our attention that the name "Directed-Evolution" was used in the completely different scientific domain of evolving enzymes to different biological functionalities. The method was pioneered by Caltech Prof. Frances Arnolds and her work was recognized with the Nobel Prize in Chemistry in 2018. We decided to keep the name "Directed-Evolution" for our work in sparsification of deep neural networks as both an acknowledgement of Prof. Arnold's work and also a nod to our "convergence". In the natural world, very different organisms might evolve similar external features, a convergence, while their internal chemistry and organization are quite different [10,11]. Analogously, we came to adopt the same name for our methods, a convergence thus, while the internal details of our technical domains are quite different.

## 2.1 Distillation context

Directed-Evolution method is used in this work in the context of distillation [7]. The distillation context is warranted by considering highly sparsified neural networks for edge computing benefit from being networks derived

| Layer (type) | Output Shape | Param # |
|---|---|---|
| flatten (Flatten) | (None, 784) | 0 |
| dense (Dense) | (None, 128) | 100480 |
| dropout (Dropout) | (None, 128) | 0 |
| dense_1 (Dense) | (None, 10) | 1290 |

Total params: 101,770
Trainable params: 101,770
Non-trainable params: 0

(a)

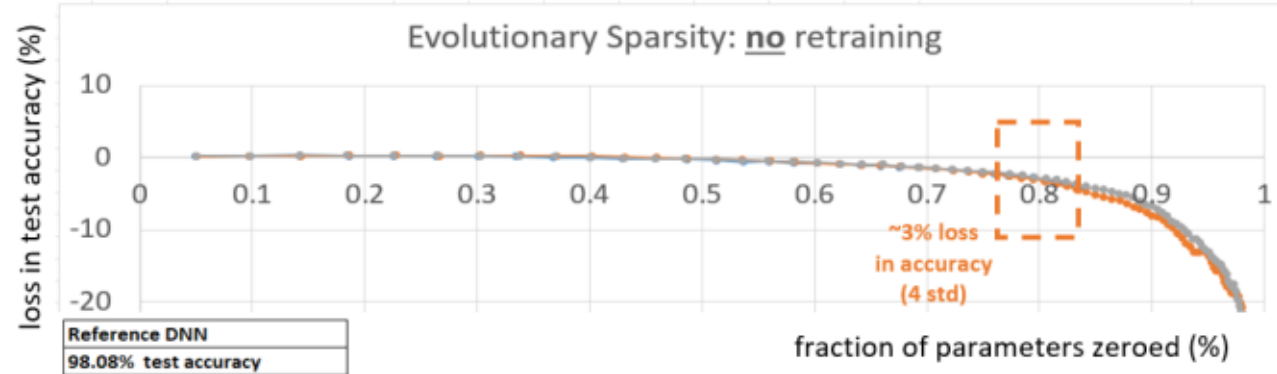

| Reference DNN | | | | | | | | | | |
|---|---|---|---|---|---|---|---|---|---|---|
| 98.08% test accuracy | | | | | | | | | | |
| 1,000 trials | | | | | | | | | | |
| zero-fraction (%) | 10 | 20 | 30 | 40 | 50 | 60 | 70 | 80 | 90 | 95 |
| mean loss of accuracy (%) | -0.1193 | -0.3884 | -0.8390 | -1.6244 | -2.9748 | -5.8080 | -11.6277 | -25.3095 | -54.1273 | -74.9472 |
| std loss of accuracy (%) | 0.0795 | 0.1372 | 0.2588 | 0.5287 | 0.9680 | 1.9476 | 3.4198 | 5.7342 | 6.9567 | 4.9432 |

(b)

| Sparsity | 80% |
|---|---|
| original accuracy | 98.08% |
| accuracy with sparsity (no retraining) | 95.13% |
| accuracy after retraining | 97.91% |
| accuracy after stochastic quantization (8bits) | 98.00% |

(c)

| Layer (type) | Output Shape | Param # |
|---|---|---|
| conv2d (Conv2D) | (None, 28, 28, 32) | 320 |
| activation | (None, 28, 28, 32) | 0 |
| batch_normalization_0 | (None, 28, 28, 32) | 128 |
| conv2d_1 | (None, 28, 28, 32) | 9248 |
| activation_1 | (None, 28, 28, 32) | 0 |
| batch_normalization_1 | (None, 28, 28, 32) | 128 |
| max_pooling2d | (None, 14, 14, 32) | 0 |
| Dropout | (None, 14, 14, 32) | 0 |
| conv2d_2 | (None, 14, 14, 64) | 18496 |
| activation_2 | (None, 14, 14, 64) | 0 |
| batch_normalization_2 | (None, 14, 14, 64) | 256 |
| conv2d_3 | (None, 14, 14, 64) | 36928 |
| activation_3 | (None, 14, 14, 64) | 0 |
| batch_normalization_3 | (None, 14, 14, 64) | 256 |
| max_pooling2d_1 | (None, 7, 7, 64) | 0 |
| dropout_1 | (None, 7, 7, 64) | 0 |
| flatten | (None, 3136) | 0 |
| dense | (None, 512) | 1606144 |
| activation_4 | (None, 512) | 0 |
| batch_normalization_4 | (None, 512) | 2048 |
| dropout_2 | (None, 512) | 0 |
| dense_1 | (None, 10) | 5130 |

Total params: 1,679,082
Trainable params: 1,677,674
Non-trainable params: 1,408

(d)

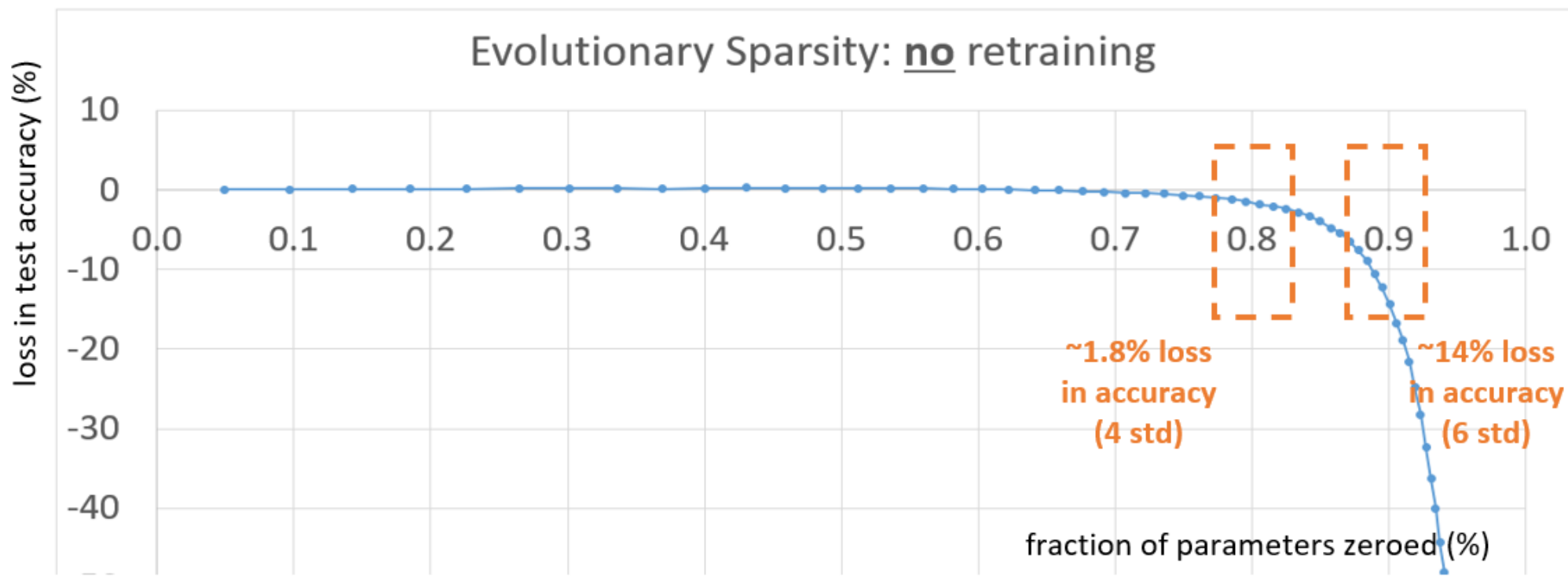

| Reference DNN | | | | | | | | | | |
|---|---|---|---|---|---|---|---|---|---|---|
| 94.37% test accuracy | | | | | | | | | | |
| 120 trials | | | | | | | | | | |
| zero-fraction (%) | 10 | 20 | 30 | 40 | 50 | 60 | 70 | 80 | 90 | 95 |
| mean loss of accuracy (%) | -0.0493 | -0.10817 | -0.15667 | -0.2385 | -0.38767 | -0.62009 | -1.14308 | -3.50225 | -38.7709 | -82.0497 |
| std loss of accuracy (%) | 0.0570 | 0.06861 | 0.0764 | 0.08918 | 0.10653 | 0.13087 | 0.15962 | 0.43799 | 3.94861 | 1.03966 |

(e)

| Sparsity | 80% | 90% |
|---|---|---|
| original accuracy | 94.37% | 94.37% |
| accuracy with sparsity (no retraining) | 92.53% | 80.04% |
| accuracy after retraining | 94.52% | 94.27% |
| accuracy after stochastic quantization (4 bits) | 94.34% | 94.18% |

(f)

Figure 3: Preliminary studies on Directed-Evolution: sparsification of shallow and deep dominant fully connected layers. (a) Network architecture used for MNIST with Directed-Evolution sparsification cycles applied with no retraining. (b) Evolution of mean and standard deviation of 1,000 random trials with candidate parameters for sparsification in each cycle. (c) Accuracy recovered with re-training and and final stochastic quantization to 8 bits produced a loss in accuracy of only 0.08% at 80% sparsity. (d) Network architecture used for Fashion-MNIST. (e) Evolution of mean and standard deviation of 120 random trials along Directed-Evolution cycles with no retraining. (e) Accuracy after retraining surpassed original accuracy at 80% sparsity. At 90% sparsity and 4 bit stochastic quantization the loss of accuracy is only 0.19%. Final network thus reaches close to 80x compression relative to the original network memory footprint.

("student" network) from a large network ("teacher" network). The teacher network is assumed to have been trained extensively on a very large dataset.

As a derived and potentially more specialized network, the student network evolves towards a highly sparsified version of the teacher network along a path of minimal departure from the predictions of the teacher network.

It should be noticed that straightforward variations on this description of the Directed-Evolution method as described are immediate. Different layers or different set of layers in a neural network may be set with different target sparsification levels. These different targets might be set a priori, or automatically determined as the cycles of Directed-Evolution method progress.

The cumulative steps of sparsification in each Directed-Evolution cycle make irrevocable decisions on which parameters are zeroed in our current studies. Once a parameter or set of parameters is zeroed, the decision is final. Since the Directed-Evolution method cycles advance by small steps on sparsity, this irrevocable nature of sparsification decisions is relatively benign. The opportunity for the student network to retrain towards minimal divergence from the teacher before the next cycle of sparsification diminish the probability of a catastrophic accumulation of bad irrevocable sparsification decisions.

## 3. PRELIMINARY STUDIES: DIRECTED-EVOLUTION AND MNIST

MNIST and Fashion-MNIST were used for initial studies in Directed-Evolution (DE) method.

A fully-connected neural network with 101k parameters shown in figure 3(a) was used for MNIST. Re-training in every cycle of Directed-Evolution (DE) was avoided to expose the effects of sparsification by itself. Even though 1,000 random trials were used in in each Directed-Evolution trial, 120 random trials were later adopted since the mean and standard deviations of the trials in every cycle were shown to be already well defined with the smaller number of trials. Every cumulative step of sparsification attempts an augment of 5% more parameters being zero in the first layer shown in fig 3(a). That produced the desired effect of sparsification advancing at almost 5% additional zeroed parameters in the initial DE cycles, and progressive smaller cumulative steps when the layer is already significantly sparsified – and random trials over all possible parameters chooses some parameter already zeroed in the layer. This makes the advance of sparsification more cautious, with smaller effective steps, as a larger percentage of the layer is already sparsificed.

Figure 3(b) shows that, with no re-training, at 80% sparsity, the cumulative sparsification shows the random trials are producing results which affect the accuracy of the network with a mean loss of 25% and standard deviation of 5.7%. The best alternative among random trials produces only 3% loss in accuracy as shown in figure 3(c). Introducing re-training at this 80% sparsification, makes the accuracy loss be only 0.17% relative to the original fully connected network.

A much larger neural network with both convolutional and fully connected layer layers totaling 1,606,144 parameters were used for Fashion-MNIST as shown in figure 3(d). As per the observation in the study of MNIST, the number of random trials in every cycle of Directed-Evolution was reduced to 120. Cumulative steps were still aimed at 5% of the total number of parameters in a layer being sparsified. This time, we focus out attention in both the 80% and 90% sparsification levels, both cases with no retraining. This time mean and standard deviation are such that the best random trial set at 80% is 4 standard deviations from the mean loss of accuracy, and at 90%, the best set for sparsification is at 6 standard deviations. Introducing re-training for the surviving parameters, the 80% sparsity case actually became 0.15% *gain* in accuracy relative to the original network, and the 90% sparsity case presented a loss in accuracy of mere 0.10%.

Event though the studies above focused in a single layer, those were the dominating layers in both neural networks considered, representing 98.7% and 95.7% for MNIST and Fashion-MNIST neural networks. Moreover, the sparsified layers were a shallow layer and a deep layer, respectively.

After recovering the accuracy with retraining, the 80% sparsity fully connected layer used for MNIST was quantized to signed 16 bits, signed 8 bits and signed 4 bits. Several techniques were used for uniformly quantization, and the best result was found for 8 bits stochastic quantization, where the round up or down decision is dependent on a probability parameterized by the distance between the quantity to be rounded and quantized values above and below. 80% sparsification and 8 bits signed quantization led to a final low of accuracy 0.08% relative to the original network.

In the Fashion-MNIST case, 80% sparsity and 4 bits stochastic quantization led to only 0.03% loss of accuracy and 90% sparsity and 4 bits stochastic quantization let to only 0.19% loss of accuracy.

In all these cases, the quantization levels were uniformly distributed from the minimum to the maximum value of parameters in a layer. Appendix B shows that better results can be achieved if a suitable distribution of quantization levels is chosen in relation to the distribution function of parameter values in a layer. There is a cost involved in those more sophisticated mapping discussed in appendix B and, depending on how constrained the resources are in the final deployment, a simple “good enough” mapping might be the best choice. It needs to be pointed out though that for quantization with very small number of bits, optimality as informed by appendix B is necessary. The optimal distribution of levels in such small number of bits quantization can use the assistance of Look-Up Tables to map binary representations to actual parameter values to be used in the neural network computations.

Additional compression can be accrued by considering a lossless encoding like Huffman coding for the surviving weights. This steps is as effective as less entropy is found in the final surviving parameters. In practice, it has been noticed an additional 10% or 20% compression might be available in specific case [6].

It appears, from this preliminary study that tracking mean and standard deviation of the random trials effect on accuracy of the cumulatively sparsified neural network is useful in gauging how far the network is itself overparameterized and hence how far sparsification can be pushed. In both examples studied, the layer being sparsified was a dense layer. In this case, a custom layer was created to be used as drop-in-place, where specific sparsification mask was included. The next session will sparsify the entire network and will incorporate sparsification masks in custom convolutional layers that are also drop-in-place layers.

## 4. DIRECTED-EVOLUTION AND YOLOv3

A pre-trained YOLOv3 neural network with all convolution layers was used in this study of Directed-Evolution sparsification. Appendix A has a description of this neural network. This network has 61,949,149 trainable parameter The largest layers are eight layers, which have 4,718,592 parameters each. The second largest layers are ten layers, which have 1,179,648 parameters each. These two sets of layers together account for a total of 49,545,216, or 80% of the total trainable parameters. Adding the third set of largest layers, seven layers (marked with green rectangles), which have 524,288 parameters per layer, leads to a total of 53,215,232 parameters, or 86% of the total trainable parameters. Because of these specific aspects, we studied Direct-Evolution method applied to (1) a single layer from the largest set of layers, (2) the whole set of largest layers, (3) the largest and second largest set of layers, (4) the three largest set of layers, and (5) the whole YOLOv3 network. In the interest of space, we will report only case (5), the use of DE for whole YOLOv3 network.

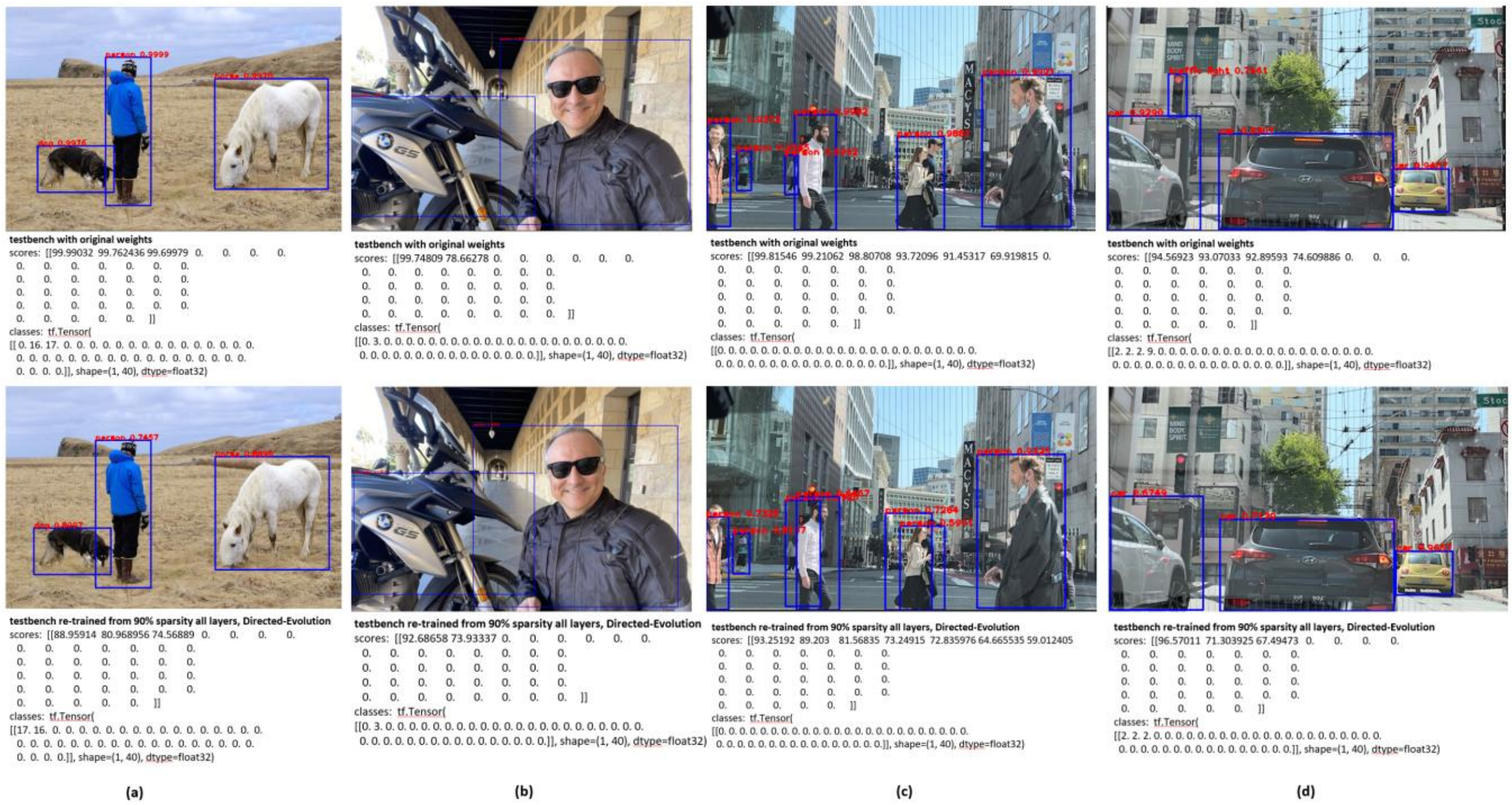

Figure 4: Directed-Evolution and YOLOv3: 90% sparsification of all layers in YOLOv3. All the 40 outputs out of 80 possible identifying the objects in an image, all the confidence outputs and all the positional outputs for segmentation produced by a pre-trained YOLOv3 output are approximated by a student YOLOv3 network sparsified to 90%. The objects identified by the teacher network with confidence above 80% are also identified by the 90% sparsified student network, albeit with a lower level of confidence. Depending on the object identified, objects that the teacher network identified with confidence level around 75% might not be identified by the student network. In (a) and (b) all the objects identified by the teacher network were identified by the student network. In (c) the student network appeared to have dropped a low confidence object while duplicating another. In (d) all objects were identified and placed in the image with the exception of the traffic light, which was identified with only 74.6% confidence by the teacher network. The dropping of objects that were identified with lower confidence levels might be improved by using a loss function different from the mean squared error (MSE) used in this study. Re-training after sparsification was limited to 1,500 epochs.

The YOLOv3 neural network also offers the opportunity to investigate the edge specialization in neural network as mentioned before. The original pre-trained YOLOv3 network was trained for classification in 80 classes over COCO dataset. This entails inferences comparisons over 100 thousand images in the distillation context used in Directed-Evolution (DE), for 120 trials every DE cycle. We decided to make an experiment and used a much smaller subset of images (less than 100 images) related to classes more relevant to home and city traffic context. In order to compensate this diminishment of training images, we use a much richer output for divergence analysis in the distillation context.

In the case of YOLOv3, we created a custom loss function based on *mean squared error* (MSE) that uses all the quantifiable outputs from the YOLOv3 network. That is, (a) the category output identification, (b) the confidence level of an identification and (c) position of the identified objects in the image. All these outputs are used and the student network been sparsified needs to produce outputs that closely mimic the teacher's (the originally trained network). All outputs in (a) and (b) take values of magnitudes between zero and one. The outputs from (c), the position, width and height for localizing the identified objects in an image needed to be normalized to the overall width and height of each image to make them of suitable magnitudes for feeding (a), (b) and (c) outputs from teacher and student networks to the MSE computation of loss.

It was noticed that advancing the sparsification at 1% steps of the parameters in a layer produced better results than using 5% steps. It might be possible that any value below 2% is adequate for most practical neural networks. At the writing of this article, only results using 5% steps are being reported.

Some representative results are shown in figure 4 for 90% sparsification using Directed-Evolution (DE) method. The divergence between teacher and student network was calculated using MSE with no priority given to any of the outputs from either network. By the very fact that classes were informed with a discrete value of zero or one, while the confidence level could have any value in the range from zero to one, the output from the student network dropped the objects for which the teacher network had confidence below around 75%. Because of that, in figure 4(d) the traffic light identification was dropped from the segmentation information by the student network. In general, all objects identified by the teacher network with confidence above 80% were also identified by the student network. These results should be improved as we move to more suitable divergence measures than MSE as used in this study.

One of the signatures of the Directed-Evolution (DE) method is shown in figure 4. Sparsification methods based solely or strongly dependent on the magnitude of parameters to decide for their zeroing tend to produce bimodal distribution for the surviving parameters. That is, the surviving parameters tend to distribute away from zero, in two concentrations of values. One concentration around a negative value and another around a positive value. Differently, in DE, the surviving parameters typically distribute as normal-looking distribution around zero value (fig.5).

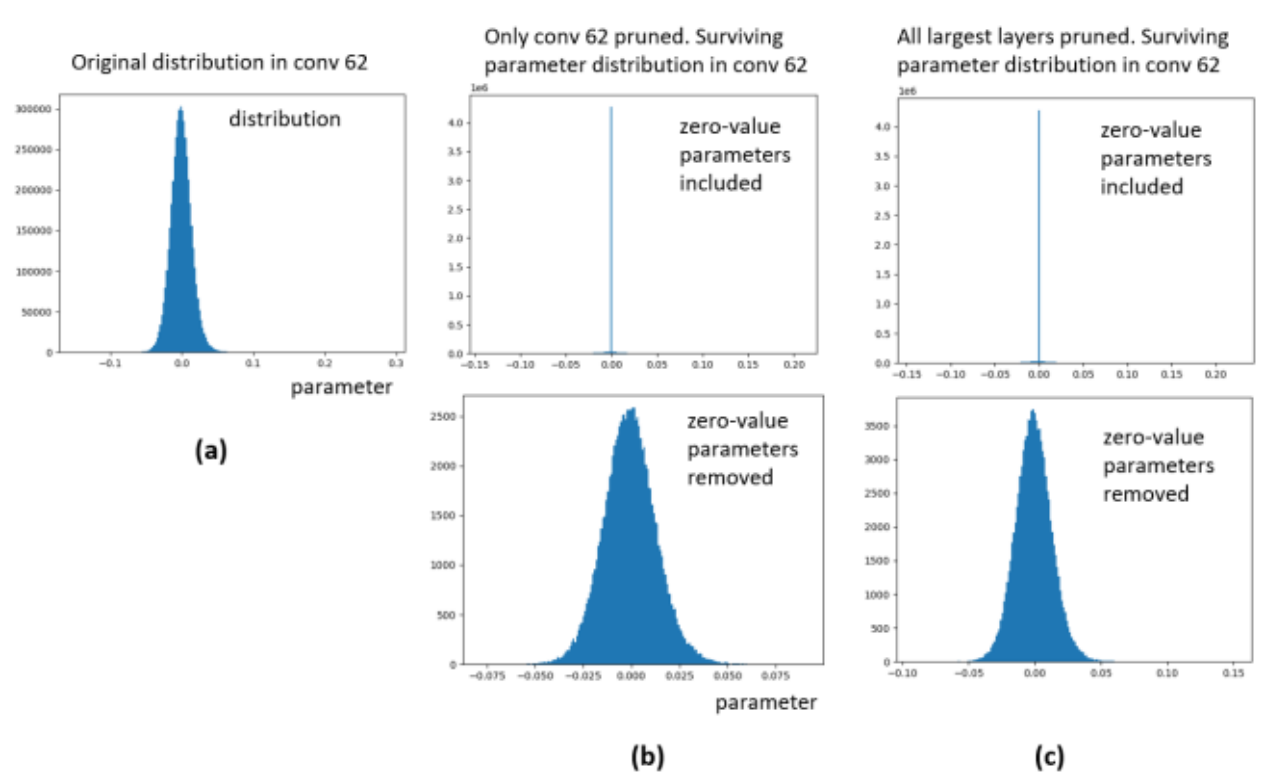

Figure 5: Directed-Evolution method tends to produce normal-looking distribution for the surviving parameters: (a) original distribution of parameters in convolution layer 62, (b) distribution of surviving parameters in convolution layer 62 when that layer was the only layer targeted for 90% sparsification, and (c) distribution of surviving parameters in convolution layer 62 when all the largest layers in YOLOv3 were target for 90% sparsification. Differently, other methods for sparsification that zeros parameters of small magnitude tend to produce bimodal distributions.

Quantization to 8-bit representation produced no change in the objects identified by the student network. Less than 1% changes in the confidence level were noticed. Optimal quantization to 4-bit also produced no change in the identified objects and most 3% lowering in confidence level in the identification of some objects.

The smallest the number of bits used in quantization, the more important is optimality in the distribution of quantization levels for each layer in the neural network. As per the mathematical development in appendix B, the separation between quantization levels need to be correspondingly smaller in direct proportion to the number of weights in that separation as expressed by the equation below:

$$(l-k)p(\frac{k+l}{2}) = (m-l)p(\frac{l+m}{2}) = cte$$

Where $l$, $k$, and $m$ are consecutive quantization intervals. The quantity of weights in the intervals *[l,k)* and *[k,m)* are informed by the distribution function values at the middle of the interval and indicated by *p((k+l)/2)* and *p((l+m)/2)*.

In the case of very low number of bits, the mapping between binary representations and actual values of parameters to be used in the neural network computations might be informed in Look-Up Tables (LUT) for any parameter distribution function found in a network layer. More details in appendix B.

## 5. FURTHER DEVELOPMENT

Directed-Evolution (DE) method, as evolution in the natural world does not guarantee the best solution for a sparsity level will always be found. It is necessary for DE to work well that a path to the best solution at a target sparsification level is reachable by small cumulative steps from lower sparsification levels. The existence of such a path is not guaranteed in the natural world either. But the requirements for DE to work in artificial neural network might be very similar to the requirements in the natural world as found by geneticist John Maynard-Smith [12]. That is, it's possible that DE worked well for the cases studied in this article and work well in general because the local minima at every sparsification level form a "dense" set of points in the space of solutions of minimal divergence between teacher and student. Dense enough to enable the DE method to navigate from 0 level of sparsity to high levels of sparsity above 90% by an unbroken path, and always be able to find a "very good" solution at each sparsity level. This paper presented the promising early successes of Directed-Evolution method from a phenomenological perspective. It's is nevertheless desirable that Directed-Evolution method be shown to be widely applicable for neural network sparsification on formal grounds next.

It is also clear that the DE method can nest criteria from other methods. For instance, in stage 2 of the DE method cycles, when the best candidate set for sparsification is determined, instead of zeroeing all the parameters in this set, additional criteria based on parameters' magnitudes and Hessian approximations can be added and only a sub-set of parameters in the best candidate set is zeroed.

## CONCLUSIONS

Directed-Evolution method using a distillation context produced results with accuracy changes of less the 1% and 4-bit representation in neural network sparsified to 90% over Fashion-MNIST. Applied to a pre-trained YOLOv3 neural network with more than 60 million parameters over COCO dataset, a student network sparsified to 90% and specialized to a smaller subset of the original classes showed that all objects identified with more than 80% confidence by the original network is also identified by the sparsified network.

Direct-Evolution makes no a priori assumption connecting directly the magnitude of a parameter and its relevance to the accuracy of the neural network. Nor make local minima curvature considerations for zeroing a parameter of the neural network. Only by direct determination of the effect of zeroing of a parameter and its effect on the network accuracy (or divergence from the original network behavior) a parameter is zeroed. Because of that, Directed-Evolution method leaves a signature in that the surviving weights presents a monomodal distribution in every layer of the network where the method is applied.

The optimal separation of quantization levels was shown to be related to the distribution function of the weights in a layer. This optimal separation becomes more important for quantizations using small number of bits. In such cases, Look-Up Tables (LUT) appear to be of convenient size to be of assistance in mapping binary representation of parameters to actual values of parameters used for computations in a neural network.

## Acknowledgments

The authors wish to thank the colleagues who provided insightful feedback on the best strategies to put to test the ideas of the Directed-Evolution method to sparsification of very large neural networks and how to make the method more computationally effective.

## References

[1] Zhu, M. & Gupta, S. To prune, or not to prune: exploring the efficacy of pruning for model compression (2017).1710.01878.

[2] He, Y., Zhang, X. & Sun, J. Channel pruning for accelerating very deep neural networks. In Proceedings of the IEEE International Conference on Computer Vision, 1389-1397 (2017).

[3] Ding, X. et al. Global sparse momentum sgd for pruning very deep neural networks. In Advances in Neural Information Processing Systems, 6382-6394 (2019).

[4] Chao Liu, Zhiyong Zhang, Dong Wang. Pruning Deep Neural Networks by Optimal Brain Damage. Interspeech 2014.

[5] Davis Blalock, Jose J G Ortiz, Jonathan Frankle. What is the state of neural network pruning? arXiv preprint arXiv:2003.03033, 2020.

[6] Song Han et al. Deep Compression: compressing deep neural networks with pruning, trained quantization and Huffman coding. ICLR 2016.

[7] Geoffrey Hinton, Oriol Vinyals, and Jeff Dean. Distilling the Knowledge in a Neural Network. *arXiv:1503.02531, 2015.*

[8] Chen, Yu-Hsin et al. "Using Dataflow to Optimize Energy Efficiency of Deep Neural Network Accelerators." IEEE Micro 37, 3 (June 2017): 12 - 21. © 2017 IEEE

[9] Syntiant Corp always-on battery power neural network accelerators. https://www.syntiant.com/

[10] Keqin Chen and Frances H Arnold. Tuning the Activity of an Enzyme for Unusual Environments: sequential random mutagenesis of subtilisin E for catalysis in dimethylformamide. PNAS vol. 90, no. 12, 1993.

[11] Frances H Arnold. Nobel Lecture 2018. https://www.nobelprize.org/prizes/chemistry/2018/arnold/lecture/

[12] John Maynard-Smith. Natural Selection and the Concept of a Protein Space. Nature vol. 225, pp. 563-564, 1970.

## A. APPENDIX

The YOLOv3 architecture we used in this paper is described in figure A.1 below.

| Layer (type) | Output Shape | Param # | Connected to |
|---|---|---|---|
| input_1 (InputLayer) | [(None, 416, 416, 3) | 0 | |
| tf.math.truediv (TFOpLambda) | (None, 416, 416, 3) | 0 | input_1[0][0] |
| conv_0 (Conv2D) | (None, 416, 416, 32) | 864 | tf.math.truediv[0][0] |
| bnorm_0 (BatchNormalization) | (None, 416, 416, 32) | 128 | conv_0[0][0] |
| leaky_0 (LeakyReLU) | (None, 416, 416, 32) | 0 | bnorm_0[0][0] |
| zero_padding2d (ZeroPadding2D) | (None, 417, 417, 32) | 0 | leaky_0[0][0] |
| conv_1 (Conv2D) | (None, 208, 208, 64) | 18432 | zero_padding2d[0][0] |
| bnorm_1 (BatchNormalization) | (None, 208, 208, 64) | 256 | conv_1[0][0] |
| leaky_1 (LeakyReLU) | (None, 208, 208, 64) | 0 | bnorm_1[0][0] |
| conv_2 (Conv2D) | (None, 208, 208, 32) | 2048 | leaky_1[0][0] |
| bnorm_2 (BatchNormalization) | (None, 208, 208, 32) | 128 | conv_2[0][0] |
| leaky_2 (LeakyReLU) | (None, 208, 208, 32) | 0 | bnorm_2[0][0] |
| conv_3 (Conv2D) | (None, 208, 208, 64) | 18432 | leaky_2[0][0] |
| bnorm_3 (BatchNormalization) | (None, 208, 208, 64) | 256 | conv_3[0][0] |
| leaky_3 (LeakyReLU) | (None, 208, 208, 64) | 0 | bnorm_3[0][0] |
| tf.__operators__.add (TFOpLambd | (None, 208, 208, 64) | 0 | leaky_3[0][0]<br>leaky_1[0][0] |
| zero_padding2d_1 (ZeroPadding2D | (None, 209, 209, 64) | 0 | tf.__operators__.add[0][0] |
| conv_5 (Conv2D) | (None, 104, 104, 128 | 73728 | zero_padding2d_1[0][0] |
| bnorm_5 (BatchNormalization) | (None, 104, 104, 128 | 512 | conv_5[0][0] |
| leaky_5 (LeakyReLU) | (None, 104, 104, 128 | 0 | bnorm_5[0][0] |
| conv_6 (Conv2D) | (None, 104, 104, 64) | 8192 | leaky_5[0][0] |
| bnorm_6 (BatchNormalization) | (None, 104, 104, 64) | 256 | conv_6[0][0] |
| leaky_6 (LeakyReLU) | (None, 104, 104, 64) | 0 | bnorm_6[0][0] |
| conv_7 (Conv2D) | (None, 104, 104, 128 | 73728 | leaky_6[0][0] |
| bnorm_7 (BatchNormalization) | (None, 104, 104, 128 | 512 | conv_7[0][0] |
| leaky_7 (LeakyReLU) | (None, 104, 104, 128 | 0 | bnorm_7[0][0] |
| tf.__operators__.add_1 (TFOpLam | (None, 104, 104, 128 | 0 | leaky_7[0][0]<br>leaky_5[0][0] |
| conv_9 (Conv2D) | (None, 104, 104, 64) | 8192 | tf.__operators__.add_1[0][0] |
| bnorm_9 (BatchNormalization) | (None, 104, 104, 64) | 256 | conv_9[0][0] |
| leaky_9 (LeakyReLU) | (None, 104, 104, 64) | 0 | bnorm_9[0][0] |
| conv_10 (Conv2D) | (None, 104, 104, 128 | 73728 | leaky_9[0][0] |
| bnorm_10 (BatchNormalization) | (None, 104, 104, 128 | 512 | conv_10[0][0] |
| leaky_10 (LeakyReLU) | (None, 104, 104, 128 | 0 | bnorm_10[0][0] |
| tf.__operators__.add_2 (TFOpLam | (None, 104, 104, 128 | 0 | leaky_10[0][0]<br>tf.__operators__.add_1[0][0] |
| zero_padding2d_2 (ZeroPadding2D | (None, 105, 105, 128 | 0 | tf.__operators__.add_2[0][0] |
| conv_12 (Conv2D) | (None, 52, 52, 256) | 294912 | zero_padding2d_2[0][0] |
| bnorm_12 (BatchNormalization) | (None, 52, 52, 256) | 1024 | conv_12[0][0] |
| leaky_12 (LeakyReLU) | (None, 52, 52, 256) | 0 | bnorm_12[0][0] |
| conv_13 (Conv2D) | (None, 52, 52, 128) | 32768 | leaky_12[0][0] |
| bnorm_13 (BatchNormalization) | (None, 52, 52, 128) | 512 | conv_13[0][0] |
| leaky_13 (LeakyReLU) | (None, 52, 52, 128) | 0 | bnorm_13[0][0] |
| conv_14 (Conv2D) | (None, 52, 52, 256) | 294912 | leaky_13[0][0] |
| bnorm_14 (BatchNormalization) | (None, 52, 52, 256) | 1024 | conv_14[0][0] |
| leaky_14 (LeakyReLU) | (None, 52, 52, 256) | 0 | bnorm_14[0][0] |
| tf.__operators__.add_3 (TFOpLam | (None, 52, 52, 256) | 0 | leaky_14[0][0]<br>leaky_12[0][0] |
| conv_16 (Conv2D) | (None, 52, 52, 128) | 32768 | tf.__operators__.add_3[0][0] |
| bnorm_16 (BatchNormalization) | (None, 52, 52, 128) | 512 | conv_16[0][0] |
| leaky_16 (LeakyReLU) | (None, 52, 52, 128) | 0 | bnorm_16[0][0] |
| conv_17 (Conv2D) | (None, 52, 52, 256) | 294912 | leaky_16[0][0] |
| bnorm_17 (BatchNormalization) | (None, 52, 52, 256) | 1024 | conv_17[0][0] |
| leaky_17 (LeakyReLU) | (None, 52, 52, 256) | 0 | bnorm_17[0][0] |

(a)

| | | | |
|---|---|---|---|
| tf.__operators__.add_4 (TFOpLam | (None, 52, 52, 256) | 0 | leaky_17[0][0]<br>tf.__operators__.add_3[0][0] |
| conv_19 (Conv2D) | (None, 52, 52, 128) | 32768 | tf.__operators__.add_4[0][0] |
| bnorm_19 (BatchNormalization) | (None, 52, 52, 128) | 512 | conv_19[0][0] |
| leaky_19 (LeakyReLU) | (None, 52, 52, 128) | 0 | bnorm_19[0][0] |
| conv_20 (Conv2D) | (None, 52, 52, 256) | 294912 | leaky_19[0][0] |
| bnorm_20 (BatchNormalization) | (None, 52, 52, 256) | 1024 | conv_20[0][0] |
| leaky_20 (LeakyReLU) | (None, 52, 52, 256) | 0 | bnorm_20[0][0] |
| tf.__operators__.add_5 (TFOpLam | (None, 52, 52, 256) | 0 | leaky_20[0][0]<br>tf.__operators__.add_4[0][0] |
| conv_22 (Conv2D) | (None, 52, 52, 128) | 32768 | tf.__operators__.add_5[0][0] |
| bnorm_22 (BatchNormalization) | (None, 52, 52, 128) | 512 | conv_22[0][0] |
| leaky_22 (LeakyReLU) | (None, 52, 52, 128) | 0 | bnorm_22[0][0] |
| conv_23 (Conv2D) | (None, 52, 52, 256) | 294912 | leaky_22[0][0] |
| bnorm_23 (BatchNormalization) | (None, 52, 52, 256) | 1024 | conv_23[0][0] |
| leaky_23 (LeakyReLU) | (None, 52, 52, 256) | 0 | bnorm_23[0][0] |
| tf.__operators__.add_6 (TFOpLam | (None, 52, 52, 256) | 0 | leaky_23[0][0]<br>tf.__operators__.add_5[0][0] |
| conv_25 (Conv2D) | (None, 52, 52, 128) | 32768 | tf.__operators__.add_6[0][0] |
| bnorm_25 (BatchNormalization) | (None, 52, 52, 128) | 512 | conv_25[0][0] |
| leaky_25 (LeakyReLU) | (None, 52, 52, 128) | 0 | bnorm_25[0][0] |
| conv_26 (Conv2D) | (None, 52, 52, 256) | 294912 | leaky_25[0][0] |
| bnorm_26 (BatchNormalization) | (None, 52, 52, 256) | 1024 | conv_26[0][0] |
| leaky_26 (LeakyReLU) | (None, 52, 52, 256) | 0 | bnorm_26[0][0] |
| tf.__operators__.add_7 (TFOpLam | (None, 52, 52, 256) | 0 | leaky_26[0][0]<br>tf.__operators__.add_6[0][0] |
| conv_28 (Conv2D) | (None, 52, 52, 128) | 32768 | tf.__operators__.add_7[0][0] |
| bnorm_28 (BatchNormalization) | (None, 52, 52, 128) | 512 | conv_28[0][0] |
| leaky_28 (LeakyReLU) | (None, 52, 52, 128) | 0 | bnorm_28[0][0] |
| conv_29 (Conv2D) | (None, 52, 52, 256) | 294912 | leaky_28[0][0] |
| bnorm_29 (BatchNormalization) | (None, 52, 52, 256) | 1024 | conv_29[0][0] |
| leaky_29 (LeakyReLU) | (None, 52, 52, 256) | 0 | bnorm_29[0][0] |
| tf.__operators__.add_8 (TFOpLam | (None, 52, 52, 256) | 0 | leaky_29[0][0]<br>tf.__operators__.add_7[0][0] |
| conv_31 (Conv2D) | (None, 52, 52, 128) | 32768 | tf.__operators__.add_8[0][0] |
| bnorm_31 (BatchNormalization) | (None, 52, 52, 128) | 512 | conv_31[0][0] |
| leaky_31 (LeakyReLU) | (None, 52, 52, 128) | 0 | bnorm_31[0][0] |
| conv_32 (Conv2D) | (None, 52, 52, 256) | 294912 | leaky_31[0][0] |
| bnorm_32 (BatchNormalization) | (None, 52, 52, 256) | 1024 | conv_32[0][0] |
| leaky_32 (LeakyReLU) | (None, 52, 52, 256) | 0 | bnorm_32[0][0] |
| tf.__operators__.add_9 (TFOpLam | (None, 52, 52, 256) | 0 | leaky_32[0][0]<br>tf.__operators__.add_8[0][0] |
| conv_34 (Conv2D) | (None, 52, 52, 128) | 32768 | tf.__operators__.add_9[0][0] |
| bnorm_34 (BatchNormalization) | (None, 52, 52, 128) | 512 | conv_34[0][0] |
| leaky_34 (LeakyReLU) | (None, 52, 52, 128) | 0 | bnorm_34[0][0] |
| conv_35 (Conv2D) | (None, 52, 52, 256) | 294912 | leaky_34[0][0] |
| bnorm_35 (BatchNormalization) | (None, 52, 52, 256) | 1024 | conv_35[0][0] |
| leaky_35 (LeakyReLU) | (None, 52, 52, 256) | 0 | bnorm_35[0][0] |
| tf.__operators__.add_10 (TFOpLa | (None, 52, 52, 256) | 0 | leaky_35[0][0]<br>tf.__operators__.add_9[0][0] |
| zero_padding2d_3 (ZeroPadding2D | (None, 53, 53, 256) | 0 | tf.__operators__.add_10[0][0] |
| conv_37 (Conv2D) | (None, 26, 26, 512) | 1179648 | zero_padding2d_3[0][0] |
| bnorm_37 (BatchNormalization) | (None, 26, 26, 512) | 2048 | conv_37[0][0] |
| leaky_37 (LeakyReLU) | (None, 26, 26, 512) | 0 | bnorm_37[0][0] |
| conv_38 (Conv2D) | (None, 26, 26, 256) | 131072 | leaky_37[0][0] |
| bnorm_38 (BatchNormalization) | (None, 26, 26, 256) | 1024 | conv_38[0][0] |

(b)

| | | | |
|---|---|---|---|
| leaky_38 (LeakyReLU) | (None, 26, 26, 256) | 0 | bnorm_38[0][0] |
| conv_39 (Conv2D) | (None, 26, 26, 512) | 1179648 | leaky_38[0][0] |
| bnorm_39 (BatchNormalization) | (None, 26, 26, 512) | 2048 | conv_39[0][0] |
| leaky_39 (LeakyReLU) | (None, 26, 26, 512) | 0 | bnorm_39[0][0] |
| tf.__operators__.add_11 (TFOpLa | (None, 26, 26, 512) | 0 | leaky_39[0][0]<br>leaky_37[0][0] |
| conv_41 (Conv2D) | (None, 26, 26, 256) | 131072 | tf.__operators__.add_11[0][0] |
| bnorm_41 (BatchNormalization) | (None, 26, 26, 256) | 1024 | conv_41[0][0] |
| leaky_41 (LeakyReLU) | (None, 26, 26, 256) | 0 | bnorm_41[0][0] |
| conv_42 (Conv2D) | (None, 26, 26, 512) | 1179648 | leaky_41[0][0] |
| bnorm_42 (BatchNormalization) | (None, 26, 26, 512) | 2048 | conv_42[0][0] |
| leaky_42 (LeakyReLU) | (None, 26, 26, 512) | 0 | bnorm_42[0][0] |
| tf.__operators__.add_12 (TFOpLa | (None, 26, 26, 512) | 0 | leaky_42[0][0]<br>tf.__operators__.add_11[0][0] |
| conv_44 (Conv2D) | (None, 26, 26, 256) | 131072 | tf.__operators__.add_12[0][0] |
| bnorm_44 (BatchNormalization) | (None, 26, 26, 256) | 1024 | conv_44[0][0] |
| leaky_44 (LeakyReLU) | (None, 26, 26, 256) | 0 | bnorm_44[0][0] |
| conv_45 (Conv2D) | (None, 26, 26, 512) | 1179648 | leaky_44[0][0] |
| bnorm_45 (BatchNormalization) | (None, 26, 26, 512) | 2048 | conv_45[0][0] |
| leaky_45 (LeakyReLU) | (None, 26, 26, 512) | 0 | bnorm_45[0][0] |
| tf.__operators__.add_13 (TFOpLa | (None, 26, 26, 512) | 0 | leaky_45[0][0]<br>tf.__operators__.add_12[0][0] |
| conv_47 (Conv2D) | (None, 26, 26, 256) | 131072 | tf.__operators__.add_13[0][0] |
| bnorm_47 (BatchNormalization) | (None, 26, 26, 256) | 1024 | conv_47[0][0] |
| leaky_47 (LeakyReLU) | (None, 26, 26, 256) | 0 | bnorm_47[0][0] |
| conv_48 (Conv2D) | (None, 26, 26, 512) | 1179648 | leaky_47[0][0] |
| bnorm_48 (BatchNormalization) | (None, 26, 26, 512) | 2048 | conv_48[0][0] |
| leaky_48 (LeakyReLU) | (None, 26, 26, 512) | 0 | bnorm_48[0][0] |
| tf.__operators__.add_14 (TFOpLa | (None, 26, 26, 512) | 0 | leaky_48[0][0]<br>tf.__operators__.add_13[0][0] |
| conv_50 (Conv2D) | (None, 26, 26, 256) | 131072 | tf.__operators__.add_14[0][0] |
| bnorm_50 (BatchNormalization) | (None, 26, 26, 256) | 1024 | conv_50[0][0] |
| leaky_50 (LeakyReLU) | (None, 26, 26, 256) | 0 | bnorm_50[0][0] |
| conv_51 (Conv2D) | (None, 26, 26, 512) | 1179648 | leaky_50[0][0] |
| bnorm_51 (BatchNormalization) | (None, 26, 26, 512) | 2048 | conv_51[0][0] |
| leaky_51 (LeakyReLU) | (None, 26, 26, 512) | 0 | bnorm_51[0][0] |
| tf.__operators__.add_15 (TFOpLa | (None, 26, 26, 512) | 0 | leaky_51[0][0]<br>tf.__operators__.add_14[0][0] |
| conv_53 (Conv2D) | (None, 26, 26, 256) | 131072 | tf.__operators__.add_15[0][0] |
| bnorm_53 (BatchNormalization) | (None, 26, 26, 256) | 1024 | conv_53[0][0] |
| leaky_53 (LeakyReLU) | (None, 26, 26, 256) | 0 | bnorm_53[0][0] |
| conv_54 (Conv2D) | (None, 26, 26, 512) | 1179648 | leaky_53[0][0] |
| bnorm_54 (BatchNormalization) | (None, 26, 26, 512) | 2048 | conv_54[0][0] |
| leaky_54 (LeakyReLU) | (None, 26, 26, 512) | 0 | bnorm_54[0][0] |
| tf.__operators__.add_16 (TFOpLa | (None, 26, 26, 512) | 0 | leaky_54[0][0]<br>tf.__operators__.add_15[0][0] |
| conv_56 (Conv2D) | (None, 26, 26, 256) | 131072 | tf.__operators__.add_16[0][0] |
| bnorm_56 (BatchNormalization) | (None, 26, 26, 256) | 1024 | conv_56[0][0] |
| leaky_56 (LeakyReLU) | (None, 26, 26, 256) | 0 | bnorm_56[0][0] |
| conv_57 (Conv2D) | (None, 26, 26, 512) | 1179648 | leaky_56[0][0] |
| bnorm_57 (BatchNormalization) | (None, 26, 26, 512) | 2048 | conv_57[0][0] |
| leaky_57 (LeakyReLU) | (None, 26, 26, 512) | 0 | bnorm_57[0][0] |
| tf.__operators__.add_17 (TFOpLa | (None, 26, 26, 512) | 0 | leaky_57[0][0]<br>tf.__operators__.add_16[0][0] |
| conv_59 (Conv2D) | (None, 26, 26, 256) | 131072 | tf.__operators__.add_17[0][0] |

(c)

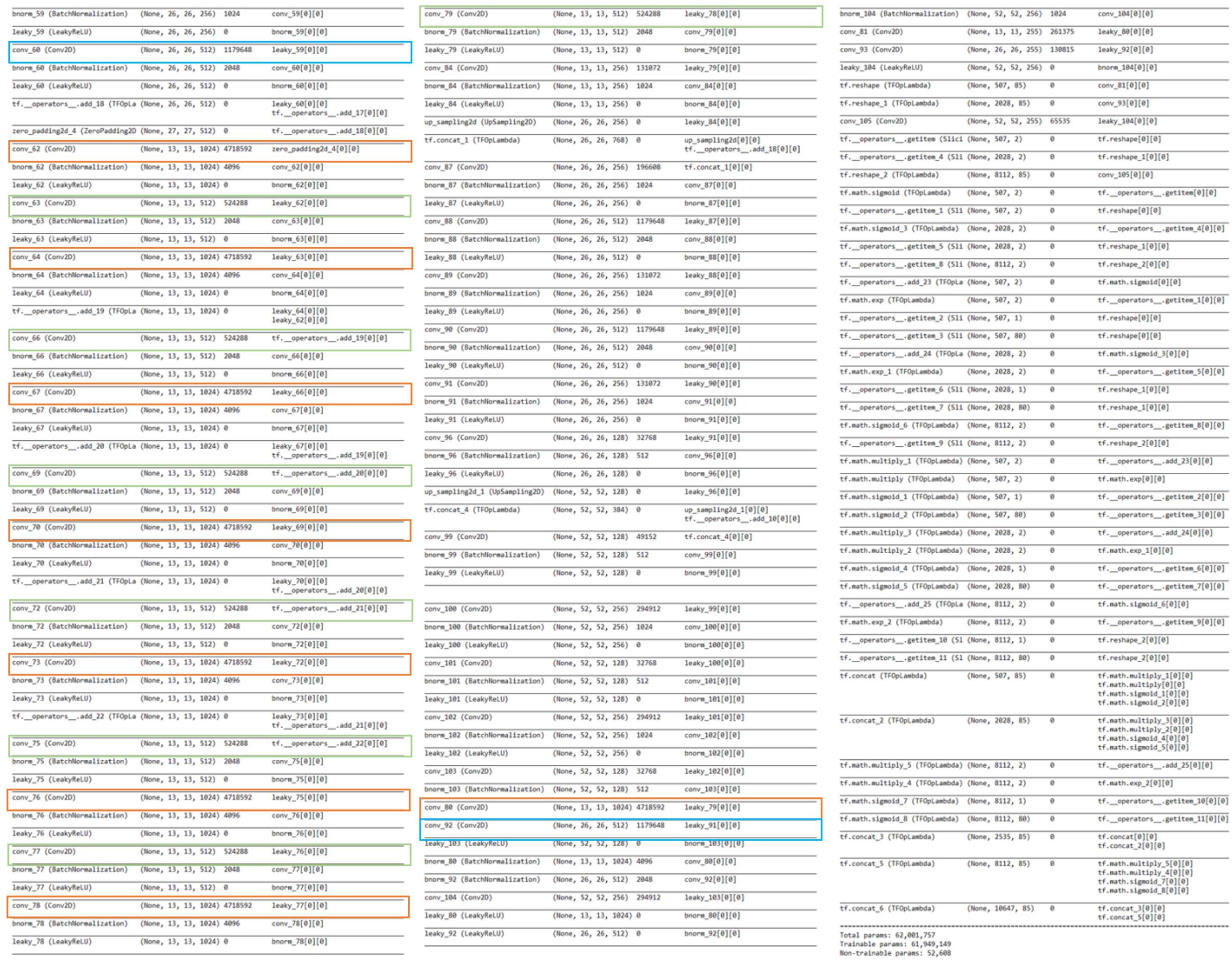

Figure A.1: YOLOv3, as used in this paper, is taken from [Reference]. The architecture has a total of 62,001,757 parameters, out of which 61,949,149 are trainable. The largest layers are eight layers (marked with orange rectangles), which have 4,718,592 parameters each. The second largest layers are ten layers (marked with blue rectangles), which have 1,179,648 parameters each. These two sets of layers together account for a total of 49,545,216, or 80% of the total trainable parameters. Adding the third set of largest layers, seven layers (marked with green rectangles), which have 524,288 parameters per layer, leads to a total of 53,215,232 parameters, or 86% of the total trainable parameters.

## B. APPENDIX

In the case where all quasi-continuous parameters are assumed equally important, optimal separation of quantization levels is established by minimizing quantization error. Figure B.1 shows a general a general distribution of weights, *w*, with distribution function *p(w)*.

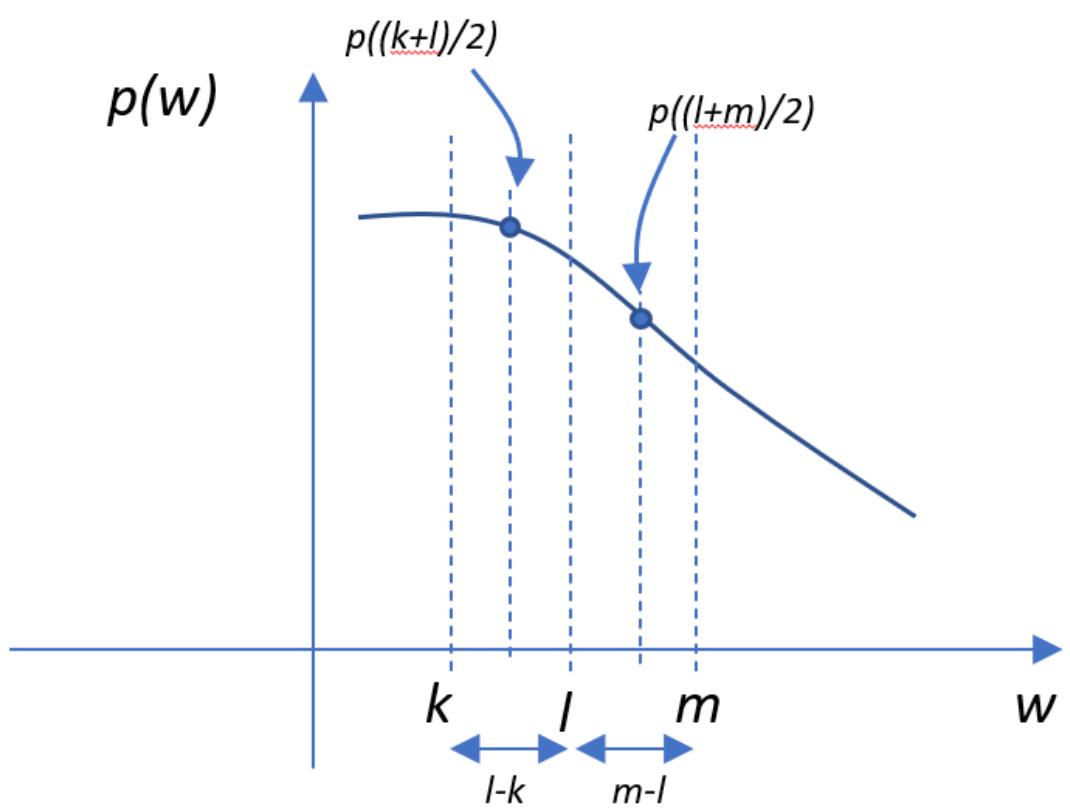

Figure B.1: a general weight distribution function, *p(w)*, and three quantization levels *k*, *l* and *m*. If the distribution of quantization levels *k*, *l* and *m* is optimal, the quantization error, $e_i$, contributed by the interval *[k,m)* is minimized. A necessary condition for optimality of *k*, *l* and *m* is that the derivative of $e_i$ relative to the *l* level position be zero valued.

The quantization process is assumed to be rounding to nearest level. This means that values of weights in the interval, for instance, [k,(k+l)/2) will be rounded to quantization level k. Values in the interval [(k+l/2,l) will be rounded to quantization level l. The quantization error contributed by the quantization levels *k*, *l*, and *m* in the interval *[k,m]* is defined as $e_i$ and expressed in equation 1.

$$e_i = \int_k^{\frac{(k+l)}{2}} (w-k)p(w)dw + \int_{\frac{(k+l)}{2}}^{l} (l-w)p(w)dw + \int_l^{\frac{(l+m)}{2}} (w-l)p(w)dw + \int_{\frac{(l+m)}{2}}^{m} (m-w)p(w)dw \qquad (1)$$

If the distribution of levels *k*, *l* and *m* is optimal, a variation $\delta l$ should produce a $\delta e_i = 0$. An expression for $\delta e_i$ is developed from equation 1 and is shown as equation 2.

$$\begin{aligned}\delta e_i = &\int_{\frac{(k+l)}{2}}^{\frac{(k+l+\delta l)}{2}} (w-k)p(w)dw \\ &+ \int_{\frac{(k+l+\delta l)}{2}}^{l+\delta l} (l+\delta l-w)p(w)dw - \int_{\frac{(k+l)}{2}}^{l} (l-w)p(w)dw \\ &+ \int_{l+\delta l}^{\frac{(l+\delta l+m)}{2}} (w-l-\delta l)p(w)dw - \int_{l}^{\frac{(l+m)}{2}} (w-l)p(w)dw \\ &- \int_{\frac{(l+m)}{2}}^{\frac{(l+\delta l+m)}{2}} (m-w)p(w)dw \qquad (2)\end{aligned}$$

Hence,

$$\delta e_i \approx \frac{l-k}{2} p(\frac{k+l}{2})\delta l/2 - \frac{m-l}{2} p(\frac{l+m}{2})\delta l/2 \qquad (3)$$

Thus, the condition

$$\frac{\delta e_i}{\delta l} = 0 \qquad (4)$$

forces

$$(l-k)p(\frac{k+l}{2}) = (m-l)p(\frac{l+m}{2}) = cte \qquad (5)$$

Equation (5) informs that the optimal distribution of quantization levels is such that separation between levels is inversely proportional to the value of the distribution function *p(w)* at the middle of the separation between levels.

A consequence of equation 5 is that if the distribution function of weights is uniform, the optimal separation of quantization levels will also be uniform as in figure B.2(a). Other arbitrary distribution functions for weights and sketches for their optimal quantization levels are sketched in figure B.2(c-e).

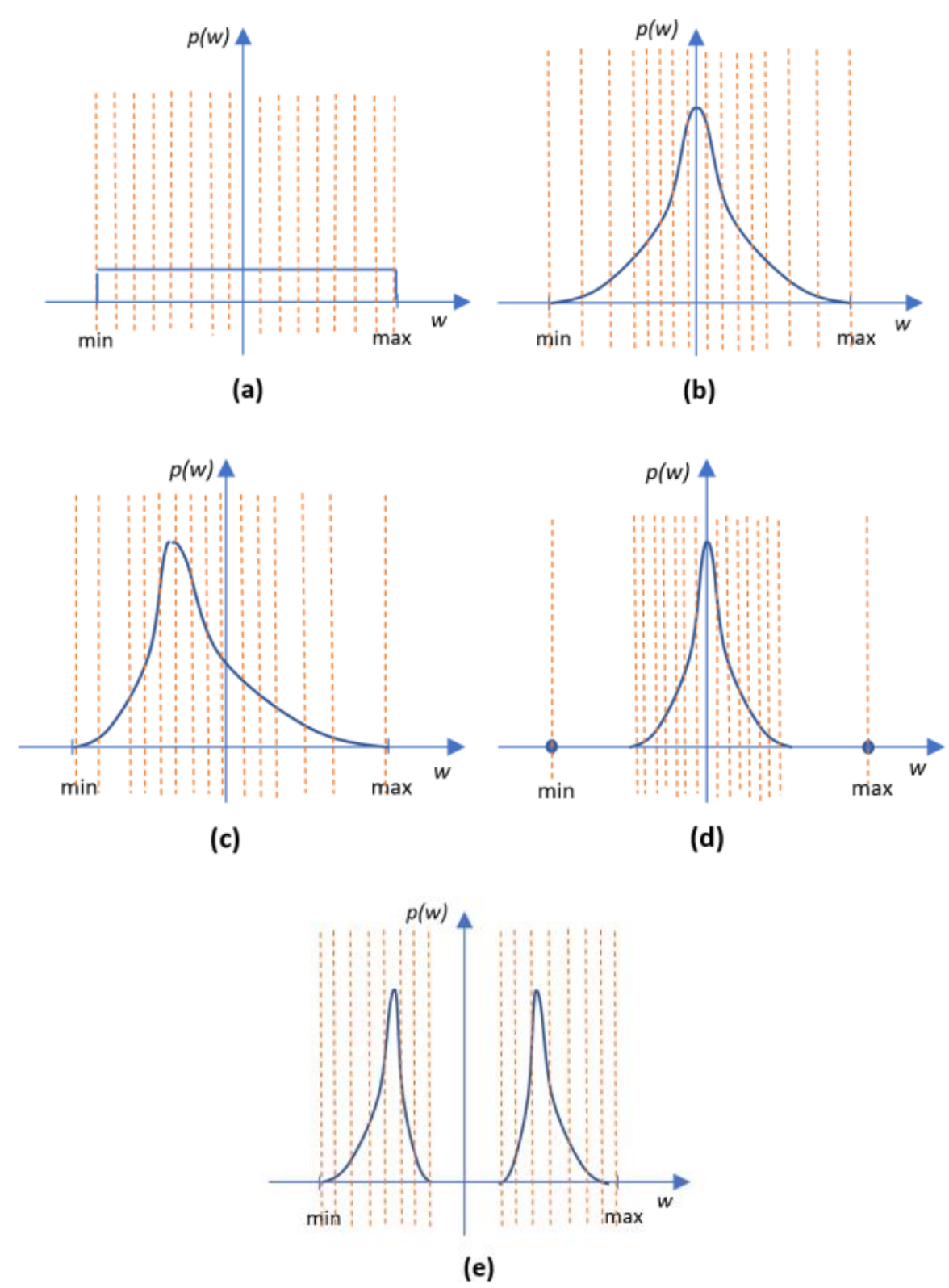

Figure B.2: Optimal distribution of quantization levels minimize quantization error. The levels are denser where the distribution function, *p(w)*, for the weights, *w*, reach highest levels. The uniform distribution of quantization levels is optimal for weights with uniform distribution function (a). The surviving weights in some pruning methods have bimodal distribution function and their optimal distribution of quantization levels is similar to the cartoon in (e).

In always-on, very low power neural network solutions, uniform quantization can take the form shown in figure B.3(a). In such, an implicit reference is made to the value zero in the weights and therefore only a scaling factor is kept for mapping weights' representation in binary and their actual values to be used in the neural network computations. In figure B.3(b) reference to the actual value of the minimum is needed for mapping. Two values, the minimum and a scaling factor, need to be kept for mapping binary representations to actual weight values in this case. B.3(b) is the better quantization scheme.

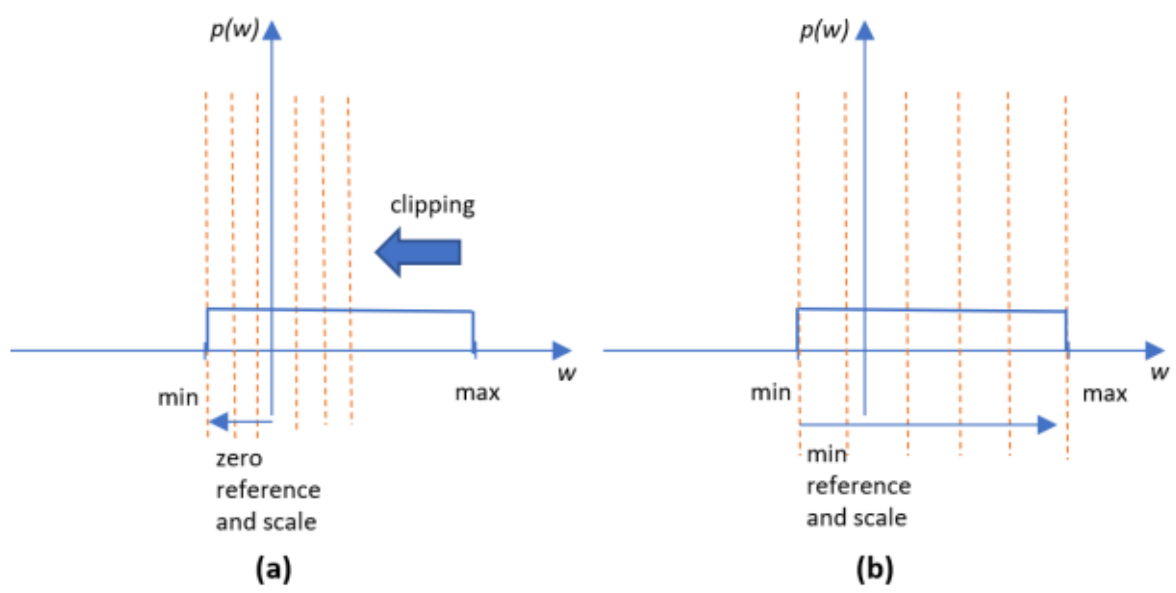

Figure B.3: Two options exist for frequently used uniform distribution of quantization levels. Only a scale factor is needed to be saved in (a): either the minimum value or the maximum value of the weights is chosen. The scale factor is used for mapping the binary representation of the weights and their actual values as used by the neural network computations. Two numbers need to be saved in (b): a scaling factor and a representation of the actual minimum value of weights. Both numbers are used to convert the binary representation of the weights to their actual values. In the case (a), some of the weights might be clipped. Case (b) is a superior use of the range of weights representation. Case (b) us the optimal distribution of quantization levels.

It should be noticed that in the case of quantization with a very small number of bits, it might be effective to use Look-Up Tables (LUT) for the optimal mapping between binary representation and actual values of weights in a layer.